\pdfoutput=1

\documentclass[11pt]{article}

\usepackage[preprint]{acl}

\usepackage{times}
\usepackage{latexsym}

\usepackage[T1]{fontenc}

\usepackage[utf8]{inputenc}

\usepackage{microtype}

\usepackage{inconsolata}

\usepackage{graphicx}
\usepackage{booktabs}
\usepackage{array}
\usepackage{amssymb}
\usepackage{amsmath} 
\usepackage{hyperref}
\usepackage{multirow} 
\usepackage{makecell}
\usepackage{xcolor}

\newcommand{\nameit}[1]{\textit{#1}}

\newcommand{\themodel}{\nameit{CuatroLLM}}

\newcommand{\thetest}{\nameit{CuatroBen}}

\newcommand{\thedata}{\nameit{TransWeb-Edu}}

\newcommand{\themodelweb}{\nameit{CuatroLLM-web}}
\newcommand{\themodelrealonly}{\nameit{WebOnlyLLM}}
\newcommand{\themodelcool}{\nameit{CuatroLLM-cool}}

\usepackage{xcolor,pifont}
\newcommand*\colourcheck[1]{%
  \expandafter\newcommand\csname #1check\endcsname{\textcolor{#1}{\ding{52}}}%
}
\newcommand*\colourcross[1]{%
  \expandafter\newcommand\csname #1cross\endcsname{\textcolor{#1}{\ding{56}}}%
}
\colourcheck{green}
\colourcheck{orange}
\colourcross{red}

\newcommand\blfootnote[1]{%
  \begingroup
  \renewcommand\thefootnote{}\footnote{#1}%
  \addtocounter{footnote}{-1}%
  \endgroup
}

\title{%
    Multilingual Pretraining Using a Large Corpus\\ Machine-Translated from a Single Source Language%
}

\author{Jiayi Wang$^{\alpha}$$^\ast$ \quad
Yao Lu$^{\alpha}$$^\ast$ \quad
Maurice Weber$^{\beta}$ \quad
Max Ryabinin$^{\beta}$ \quad
Yihong Chen$^{\alpha}$  \\
\textbf{Raphael Tang}$^{\gamma}$  \quad
\textbf{Pontus Stenetorp}$^{\alpha, \delta}$ \\
$^{\alpha}$Centre for Artificial Intelligence, University College London \quad \\
$^{\beta}$Together AI \quad $^{\gamma}$University of Waterloo \quad \\
$^{\delta}$Research and Development Center for Large Language Models, National Institute of Informatics\\
\texttt{\{jiaywang,yao.lu,yihong.chen,p.stenetorp\}@cs.ucl.ac.uk} \\
\texttt{\{webermaug,mryabinin0,tang.raphael\}@gmail.com}
}

\begin{document}
\maketitle
\blfootnote{$^\ast$ Both authors contribute equally to this work.}

\begin{abstract}
English, as a very high-resource language, enables the pretraining of high-quality large language models~(LLMs). 
The same cannot be said for most other languages, 
as leading LLMs still underperform for non-English languages, 
likely due to a gap in the quality and diversity of the available multilingual pretraining corpora.
In this work, we find that machine-translated text from a single high-quality source language can contribute significantly to the pretraining of multilingual LLMs.
We translate \nameit{FineWeb-Edu}, a high-quality English web dataset, into French, German, and Spanish, resulting in a final 300B-token dataset, which we call \thedata{},
%
and pretrain a 1.3B-parameter model, \themodel{}, from scratch on this dataset.
Across five non-English reasoning tasks, we show that \themodel{} matches or outperforms state-of-the-art multilingual models trained using closed data, such as \nameit{Llama3.2} and \nameit{Gemma2}, despite using an order of magnitude less data, such as about 6\% of the tokens used for \nameit{Llama3.2}'s training.
We further demonstrate that with additional domain-specific pretraining, amounting to less than 1\% of \thedata{}, \themodel{} surpasses the state of the art in multilingual reasoning.
To promote reproducibility, we release our corpus, models, and training pipeline under open licenses at \href{https://hf.co/britllm/CuatroLLM}{\texttt{hf.co/britllm/CuatroLLM}}.

%
%
\end{abstract}

\section{Introduction}

Large language models~(LLMs) are a central component of today's natural language processing systems due to their ability to solve a broad range of problems with little to no supervision~\citep{brown2020language,achiam2023gpt,team2023gemini, dubey2024llama}.
The success of these models is often attributed to the large-scale, high-quality datasets that they are pretrained on.
However, state-of-the-art LLMs often exhibit lower performance for languages other than English.

A common explanation for this phenomenon is the prevalence of high-quality English data in training corpora of most advanced models~\cite{penedo2023refinedweb}.
For example, in the pretraining corpus of BLOOM~\cite{le2023bloom}, the amount of English data is two and three times greater than that of French and Spanish, respectively.
In view of this perceived imbalance, works like \nameit{CroissantLLM}~\cite{faysse2024croissantllm} attempt to create multilingual models with a balanced distribution of languages during pretraining.

However, the performance gap of such approaches remains non-negligible.
We conjecture that this is due to the nature of web data, where English exhibits higher diversity and quality compared to other languages~\cite{joshi2020state,kreutzer2022quality}. 
%
Simply scaling up the size is unlikely to create a \textit{semantically} balanced multilingual dataset.
Recent efforts to address this include \citet{together2023redpajama}, which proposes to collect terabyte-scale, high-quality multilingual data and mix it with English data for pretraining.
Nevertheless, current state-of-the-art models such as \nameit{Llama3.2}~\cite{metaLlama32} still lag significantly behind even for high-resource, non-English languages, such as French, German, and Spanish.

Promisingly, \citet{etxaniz-etal-2024-multilingual} demonstrates that translating non-English queries into English at inference time often leads to consistent improvement for many reasoning tasks.
However, it is not clear whether such findings holds at pretraining time.
One may thus ask if it is possible to pretrain a performant multilingual LLM by translating a high-quality English pretraining corpus to other languages.

In this paper, we explore this research question and present a scalable method for creating a balanced multilingual pretraining dataset by translating an English pretraining corpus.
Concretely, we translate \nameit{FineWeb-Edu}~\citep{lozhkov2024fineweb-edu} into multiple languages, namely French, German, and Spanish, using \nameit{Mistral-7B-Instruct}~\cite{jiang2023mistral}, yielding a balanced multilingual pretraining corpus~(\thedata) of approximately $300$B tokens. 
Empirically, we show that high-resource languages such as French, German, and Spanish benefit from the translated corpus. 
Using \textit{solely} this translated data for pretraining, our model, \themodel, matches the performance of previous state-of-the-art multilingual language models such as \nameit{Gemma2}~\citep{team2024gemma}, \nameit{Llama3.2}~\citep{dubey2024llama}, \nameit{Qwen2}~\citep{yang2024qwen2}, and \nameit{EuroLLM}~\citep{martins2024eurollm}.

In summary, our contributions are as follows:
\begin{enumerate}
    \item To our knowledge, we are the first to translate a high-quality, pretraining-scale English corpus into multiple languages, creating a $300$B-token dataset: \thedata{}.
    \item We pretrain a $1.3$B-parameter model, \themodel{}, from scratch using \thedata{}. Despite using an order of magnitude less data (such as about 6\% of the tokens used for \nameit{Llama3.2}'s training), 
    \themodel{} matches or outperforms state-of-the-art multilingual models trained on closed data, including \nameit{Gemma2}, \nameit{Llama3.2}, \nameit{Qwen2}, and \nameit{EuroLLM}, across five non-English reasoning tasks.
    \item We release our corpus, models, and training pipeline under open licenses to promote reproducibility in multilingual NLP research.
\end{enumerate}

\section{Methodology}
This section details our methodology for constructing and utilizing machine-translated data for multilingual pretraining. On a high level, our approach can be described as follows:
\begin{enumerate}
    \item We select a high-quality, single-language \textit{source} pretraining dataset.
    \item We translate the documents from that dataset to \textit{target} languages using a \textit{translation model}.
    \item We train a language model from scratch on the resulting multilingual data mixture.
\end{enumerate}

\subsection{\thedata: the translation data}
\label{sec:trans_data}
LLMs are predominantly trained on document-level text data, as exemplified by the \nameit{LlaMA}~\citep{dubey2024llama} and \nameit{Gemma}~\citep{team2024gemma} model families. Following this convention, our translation data for pretraining is also structured at the document level. Translation data typically comprises three key components: source data, translation model, and a method for composing document-level translations.
This sub-section will elaborate on these three critical factors.

\paragraph{Source data}
The performance of an LLM is tied to the quality of its pretraining dataset. English, widely recognized as a very high-resource language, is likely to possess the greatest diversity and breadth of knowledge~\citep{joshi2020state, kreutzer2022quality}. Given this rich linguistic landscape, a high-quality English web dataset is potentially the ideal source data.

The \nameit{FineWeb-Edu} dataset~\citep{lozhkov2024fineweb-edu}, a subset of \nameit{FineWeb}~\citep{penedo2024finewebdatasetsdecantingweb}, has demonstrated its quality and efficacy in the development of LLMs. This dataset, constructed using scalable automated high-quality annotations for educational value, has been instrumental in training both English-centric models like \nameit{GPT-2}~\citep{Karpathy2022,Karpathy2024} and multilingual models such as \nameit{EuroLLM}~\citep{martins2024eurollm}, and thus emerges as a suitable candidate for our source dataset. It originally consists $1.3$ trillions tokens of educational contents, and our focus is on the sample-$100$BT subset,\footnote{\href{https://huggingface.co/datasets/HuggingFaceFW/fineweb-edu}{\texttt{hf.co/datasets/HuggingFaceFW/fineweb-edu}}} which contains about $100$ billion \nameit{GPT-2} tokens randomly sampled from the whole dataset. 
We sample around $64$ billion tokens from this subset as our source data in English.

\paragraph{Translation model}
\citet{dubey2024llama} demonstrated the potential of the Mistral model \citep{jiang2023mistral} for multilingual Natural Language Processing~(NLP) tasks. Furthermore, \citet{moslem2023fine} and \citet{kocmi2024preliminary} have highlighted its capabilities for machine translation tasks. Given these promising results, we employ \nameit{Mistral-7B-Instruct-v0.1}\footnote{\href{https://hf.co/mistralai/Mistral-7B-Instruct-v0.1}{\texttt{hf.co/mistralai/Mistral-7B-Instruct-v0.1}}} as our translation model. However, its efficacy when prompted for document-level translation, particularly with long-context English source documents, has not yet been verified. 

A recent related work by \citet{maini2024rephrasing} has empirically demonstrated that prompting an LLM to rephrase more than $300$ tokens could lead to information loss when rephrasing web data. 
%
Following their setup, we segment the English source documents from the sample-$100$BT subset of \nameit{FineWeb-Edu} into shorter pieces, prompt Mistral to translate these segments sequentially, and subsequently reconstruct the whole translated document by concatenating the translated segments.\footnote{The detailed translation pipeline is shown in Appendix~\ref{sec:chunking}.}

\begin{figure}[!t]
\includegraphics[width=1.0\columnwidth]{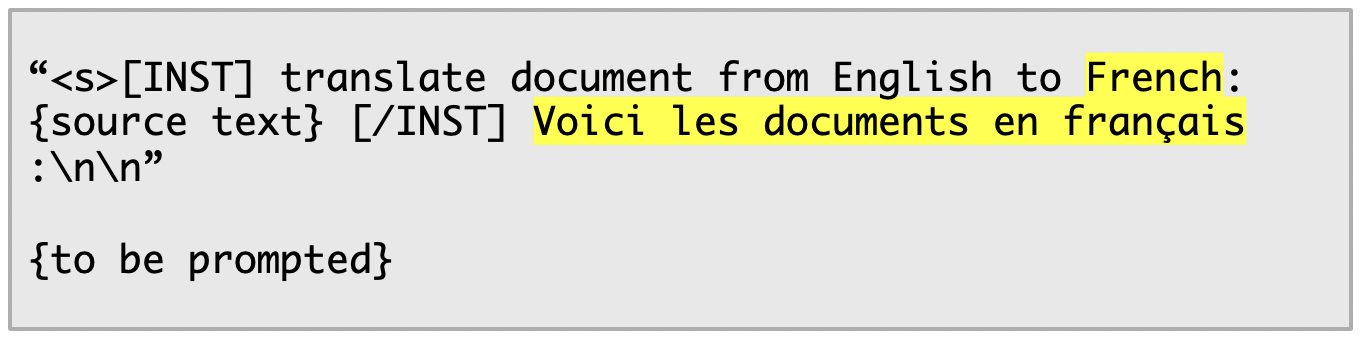}
\caption{Chat template used for prompting \nameit{Mistral-7B-Instruct-v0.1} for English-French translation.}
\label{fig:prompt_template}
\end{figure}

Adhering to the instruction format specified\footnote{\href{https://hf.co/mistralai/Mistral-7B-Instruct-v0.1}{\texttt{hf.co/mistralai/Mistral-7B-Instruct-v0.1}}} for \nameit{Mistral-7B-Instruct}, the chat template employed to prompt Mistral model for translation~(using English-French as an example) is illustrated in Figure~\ref{fig:prompt_template}.\footnote{The highlighted portions in the template are adjusted according to the target language.} 
To maintain translation integrity, any sentence not fully translated to a terminal punctuation is omitted, based on the NLTK sentence tokenizer~\citep{bird2009natural}.

\begin{table}[!t]
\small
\centering
\begin{tabular}{l|cc}
\toprule
Language & \multicolumn{1}{c}{Tokens (B)} & \multicolumn{1}{c}{Avg. Doc Length (tokens)} \\
\midrule
English  & 63.4 & 1,171.5 \\
French   & 76.3 & 1,408.7 \\
German   & 73.9 & 1,365.4 \\
Spanish  & 72.9 & 1,383.3 \\
\midrule
Total &  286.5 & 1,338.6\\
\bottomrule
\end{tabular}

\caption{\thedata{} Statistics.}
\label{tab:data_statistics}
\end{table}

\paragraph{\thedata}
We translate English documents from \nameit{FineWeb-Edu} \citep{lozhkov2024fineweb-edu} into three major European languages: French, German, and Spanish. We call this multiway parallel translated dataset as \thedata. To optimize memory efficiency and accelerate the inference process of \nameit{Mistral-7B-Instruct-v0.1}, we employ \nameit{vLLM} \citep{kwon2023efficient}, a library specifically designed for efficient large language model inference and serving. Using this setup, we translate approximately $54$ million English documents into the three target languages by prompting \nameit{Mistral-7B-Instruct-v0.1}. Table~\ref{tab:data_statistics} presents the statistics of the original English data and the translated French, German, and Spanish components of \thedata. Leveraging \nameit{vLLM}'s efficiency, we estimate the total computational cost to be approximately $6.03 \times 10^{22}$ FLOPs.  To the best of our knowledge, \thedata{} represents the largest multiway parallel, document-level multilingual pretraining dataset currently available.

\begin{table*}[!t]
\small
\centering
\resizebox{1.0\textwidth}{!}{%
\begin{tabular}{lcccccc}
\toprule
\multirow{2}{*}{Model} & \multirow{2}{*}{\# Param.} & \multirow{2}{*}{Corpus} & Corpus & Training & Data & \multirow{2}{*}{Languages} \\
& & & Size & Tokens & Avail.& \\
\midrule
\multicolumn{7}{l}{\textit{Monolingual LLMs}} \\
\href{https://huggingface.co/asi/gpt-fr-cased-base}{GPT-fr} & 1B & Wiki, OpenSubtitle, Gutenberg, Common Crawl& 3.11B & 16.3B & \redcross & en,fr\\
\href{https://huggingface.co/TinyLlama/TinyLlama_v1.1}{TinyLlama} & 1.1B & SlimPajama,StarCoder training data & 1T & 3T & \greencheck & primarily en \\
\href{https://huggingface.co/EleutherAI/pythia-1.4b}{Pythia} & 1.4B & The Pile & 207B & 300B & \greencheck & primarily en \\
\midrule
\multicolumn{7}{l}{\textit{Multilingual LLMs}} \\
\href{https://huggingface.co/ai-forever/mGPT}{mGPT} & 1.3B & mC4,Wiki & 488B & 440B & \redcross & {\makecell{en, fr, es, de,\\and others (61 total)}}\\
\href{https://huggingface.co/bigscience/bloomz-1b1}{BLOOM} & 1.1B & {\makecell{BigScience Catalogue, Common Crawl, \\Github Code, OSCAR}} & 350B & 366B & \redcross & {\makecell{en, fr, es, de,\\and others (46 total)}} \\
\href{https://huggingface.co/croissantllm/CroissantLLMBase}{CroissantLLM} & 1.3B & Croissant & 1T & 3T& \greencheck & en, fr\\
\href{https://huggingface.co/meta-llama/Llama-3.2-1B}{Llama-3.2} & 1B &-&- & 9T & \redcross & {\makecell{en, fr, es, de,\\and others (8 total)}}\\
\href{https://huggingface.co/utter-project/EuroLLM-1.7B}{EuroLLM} & 1.7B & {\makecell{mC4, Parallel Data, Code/Math, Wiki, \\ArXiv, Books, Apollo, Annealing Data}} & - & 4T & \redcross & {\makecell{en, fr, es, de,\\and others (35 total)}}\\
\href{https://huggingface.co/Qwen/Qwen2-1.5B}{Qwen2} & 1.5B &- & - & 7T & \redcross & {\makecell{en, fr, es, de,\\and others (30 total)}}\\
\href{https://huggingface.co/google/gemma-2b}{Gemma2} & 2B & web documents, code, and science articles & - & 2T & \redcross & {\makecell{en, fr, es, de,\\and others}} \\
\href{https://huggingface.co/britllm/CuatroLLM}{\themodel{}} (Ours) & 1.3B & \thedata & 287B & 515B & \greencheck & en, fr, de, es \\
\bottomrule
\end{tabular}
}
\caption{Overview of pretraining data among LLMs.}
\label{tab:model_comparison}
\end{table*}
\subsection{Multilingual pretraining}
In this section, we will describe the technical details of pretraining multilingual language models from scratch using our \thedata{} dataset.

\paragraph{Model architecture and hyper-parameters} \themodel{}'s architecture and hyperparameter selection (detailed in Appendix Section~\ref{sec:appendix-model-arch-params}) are based on lessons from the \nameit{Llama} family of models and other open-source efforts~\cite{Karpathy2024} to reproduce \nameit{GPT-2}. 
The model has a total of $1.3$B parameters.
Similar to \nameit{GPT-2} reproduction efforts, we use a constant learning rate of $6\times 10^{-4}$ and a sequence length of up to $2048$ tokens, with a batch size of $1024$ per iteration, resulting in approximately $2$ million tokens processed per iteration.

\paragraph{Tokenization}  In this work, we focus on four high-resource languages: English, French, Spanish, and German. Following previous attempt~\citep{alves2024tower} at extending multilingual capabilities of \nameit{Llama2} models~\citep{touvron2023llama2}, we found that the \nameit{Llama2} tokenizer provides very high-quality coverage for English, French, Spanish, and German. Therefore, for simplicity and effectiveness, we use the \nameit{Llama2} tokenizer. 


\paragraph{Pretraining data} Although \thedata{} is one of the largest multiway parallel, document-level multilingual datasets available, we do not intend to create aligned translation pairs for pretraining. Instead, we adopt a monolingual pretraining setup, where documents are randomly fetched from the corpus. This approach means that the probability of the same document appearing in different languages within a single batch is very low. For clarity, we illustrate our pretraining sequence in Table~\ref{tab:training-example}.
\begin{table}[!t]
\small
\centering
\resizebox{1.0\columnwidth}{!}{%
\begin{tabular}{l}
\toprule
\textbf{GPT2-style Training Sequence} \\
\midrule
\texttt{<random-en-doc><eos><random-fr-doc><eos>....<random-de-doc>} \\
\bottomrule
\end{tabular}
}
\caption{Illustration of our pretraining sequence containing multiple non-parallel documents.}
\label{tab:training-example}
\end{table}

\paragraph{Framework and training} \themodel{} is trained using the \nameit{Megatron-LM} framework~\citep{shoeybi2019megatron} with an accelerated attention implementation~\citep{dao2023flashattention}.  
The training is done on the UK Isambard-AI NVIDIA GH200 cluster~\citep{mcintosh2024isambard} for $3,187$ GPU hours.
Our pretraining dataset \thedata{} is created on a balanced basis, so we do not perform any up-sampling for specific languages. Our pretraining over \thedata{} processes approximately $515$B tokens, which is nearly two epochs of the pretraining data. Similar to the observation of \citet{muennighoff2024datascaling}, we observe no degradation in performance over the validation set during this process.

\section{Experiments}
This section presents our evaluation of model performance across several multilingual benchmarks.

\subsection{\thetest{}}
Our evaluation framework combines general English benchmarks with extended assessments focused on French, German, and Spanish. This approach enables the evaluation of model capabilities across these four languages.
We call this collection of benchmarks as \textbf{\thetest{}}. All benchmarks in \thetest{} are publicly available and open-source, ensuring transparency and reproducibility of our results.\footnote{We conduct all evaluations using~\url{https://github.com/EleutherAI/lm-evaluation-harness}.} Specifically, \thetest{} has the following complex reasoning tasks:

\textbf{Hellaswag}~\citep{zellers2019hellaswag}: Common-sense reasoning requiring models to predict contextually appropriate sentence endings.

\textbf{ARC} \citep{clark2018think}: $7,787$ grade-school level, multiple-choice science questions to assess QA capabilities.

\textbf{TruthfulQA}~\citep{lin2021truthfulqa}: Evaluates the truthfulness of LLMs in generating answers across health, law, finance, and politics domains.

\textbf{PAWS-X} \citep{yang-etal-2019-paws}: A cross-lingual adversarial dataset for paraphrase identification, consisting of translated pairs originally sourced from English Wikipedia and Quora.

\textbf{XNLI}~\citep{conneau2018xnli}: An extension of the Multi-Genre NLI \citep{williams-etal-2018-broad} to $15$ languages, assessing textual entailment prediction.

For Hellaswag, ARC, and TruthfulQA, which are originally English-only benchmarks, we use the translations from ChatGPT\footnote{\url{https://openai.com/index/chatgpt/}} provided by \citet{lai2023okapi}.\footnote{For French, German and Spanish, only the ARC-Challenge (ARC-C) sets are available.} Additionally, for English-specific evaluation, we include \textbf{PIQA} \citep{Bisk2020} to assess physical commonsense reasoning and \textbf{SciQ} \citep{SciQ} for multiple-choice science exam questions covering Physics, Chemistry, and Biology. All benchmarks in this set are evaluated using a standard few-shot~($5$-shot) setting, with results reported as accuracy.

\begin{table}[!t]
\centering
\resizebox{\columnwidth}{!}{%
\begin{tabular}{lccccc|c}
\toprule
 \textbf{Model} & \textbf{ARC-C} & \textbf{Hellaswag} & \textbf{PAWS} & \textbf{TruthfulQA} & \textbf{XNLI} & \textbf{Mean} \\
\midrule
\multicolumn{7}{l}{\textit{Monolingual LLMs}} \\
GPT-fr & 19.85 & 29.32 & 52.55 & 24.27 & 37.91 & 32.78 \\
Pythia & 20.44 & 29.64 & 54.55 & 27.45 & 43.86 & 35.19 \\
TinyLlama & 24.64 & 32.67 & 50.95 & 29.22 & 42.49 & 35.99 \\
\midrule
\multicolumn{7}{l}{\textit{Multilingual LLMs}} \\
mGPT & 20.79 & 27.14 & 53.60 & 22.49 & 40.56 & 32.92 \\
BLOOM & 23.18 & 33.74 & 51.00 & 26.30 & 45.86 & 36.02 \\
CroissantLLM & 25.75 & 39.69 & 51.55 & 23.38 & 44.22 & 36.92 \\
Llama3.2 & 27.46 & 36.02 & 52.20 & 28.46 & 43.94 & 37.62 \\
Qwen2 & 29.60 & 38.78 & 48.15 & 29.10 & 44.62 & 38.05 \\
\underline{EuroLLM} & 32.34 & 40.27 & 51.90 & 27.45 & 45.14 & 39.42 \\
\underline{Gemma2} & 35.76 & 39.70 & 48.55 & 28.08 & 47.35 & 39.89 \\

\midrule
\underline{\themodel} & 33.53 & 38.00 & 52.10 & 26.43 & 42.61 & 38.53 \\
\bottomrule
\end{tabular}
}
\caption{\thetest-French Performance.}
\label{tab:cuatrobench_french_results}
\end{table}

\begin{table}[!t]
\centering
\resizebox{\columnwidth}{!}{%
\begin{tabular}{lccccc|c}
\toprule
\textbf{Model} & \textbf{ARC-C} & \textbf{Hellaswag} & \textbf{PAWS} & \textbf{TruthfulQA} & \textbf{XNLI} & \textbf{Mean} \\
\midrule
\multicolumn{7}{l}{\textit{Monolingual LLMs}} \\
Pythia & 19.59 & 28.54 & 47.90 & 27.54 & 41.53 & 33.02 \\
TinyLlama & 21.56 & 30.66 & 49.50 & 24.75 & 40.52 & 33.40 \\
\midrule
\multicolumn{7}{l}{\textit{Multilingual LLMs}} \\
BLOOM & 20.10 & 27.18 & 49.20 & 26.78 & 37.67 & 32.19 \\
mGPT & 19.42 & 27.69 & 51.55 & 21.19 & 40.56 & 32.08 \\
Qwen2 & 26.69 & 34.77 & 43.40 & 28.55 & 42.77 & 35.24 \\
Llama3.2 & 26.78 & 34.11 & 47.95 & 27.54 & 44.22 & 36.12 \\
\underline{Gemma2} & 31.14 & 37.31 & 42.55 & 26.65 & 46.63 & 36.86 \\
\underline{EuroLLM} & 29.43 & 37.47 & 46.00 & 28.68 & 46.10 & 37.54 \\
\midrule
\underline{\themodel} & 31.39 & 35.70 & 51.35 & 26.90 & 42.57 & 37.58 \\
\bottomrule
\end{tabular}
}
\caption{\thetest-German Performance.}
\label{tab:cuatrobench_german_results}
\end{table}

\begin{table}[!t]
\centering
\resizebox{\columnwidth}{!}{%
\begin{tabular}{lccccc|c}
\toprule
\textbf{Model} & \textbf{ARC-C} & \textbf{Hellaswag} & \textbf{PAWS} & \textbf{TruthfulQA} & \textbf{XNLI} & \textbf{Mean} \\
\midrule
\multicolumn{7}{l}{\textit{Monolingual LLMs}} \\
Pythia & 22.14 & 30.26 & 50.15 & 29.40 & 41.45 & 34.68 \\
TinyLlama & 24.10 & 33.37 & 49.15 & 29.40 & 42.73 & 35.75 \\
\midrule
\multicolumn{7}{l}{\textit{Multilingual LLMs}} \\
mGPT & 20.51 & 28.42 & 53.15 & 21.80 & 39.88 & 32.75 \\
BLOOM & 24.96 & 34.49 & 51.15 & 26.62 & 43.86 & 36.22 \\
Llama3.2 & 28.72 & 37.13 & 51.55 & 27.50 & 43.01 & 37.58 \\
Qwen2 & 30.51 & 39.03 & 44.05 & 31.43 & 43.09 & 37.62 \\
\underline{Gemma2} & 36.15 & 41.69 & 44.00 & 28.01 & 44.34 & 38.84 \\
\underline{EuroLLM} & 32.91 & 41.00 & 49.75 & 27.25 & 43.98 & 38.98 \\
\midrule
\underline{\themodel} & 32.99 & 38.66 & 50.00 & 27.88 & 42.97 & 38.50 \\
\bottomrule
\end{tabular}
}
\caption{\thetest-Spanish Performance.}
\label{tab:cuatrobench_spanish_results}
\end{table}

\begin{table}[!t]
\centering
\resizebox{\columnwidth}{!}{%
\begin{tabular}{lcccccccc|c}
\toprule
\textbf{Model} & \textbf{ARC-C} & \textbf{ARC-E} & \textbf{Hellaswag} & \textbf{PAWS} & \textbf{PIQA} & \textbf{SciQ} & \textbf{TruthfulQA} & \textbf{XNLI} & \textbf{Mean} \\
\midrule
\multicolumn{10}{l}{\textit{Monolingual LLMs}} \\
Pythia & 27.99 & 64.23 & 40.49 & 46.65 & 71.06 & 91.80 & 22.77 & 48.59 & 51.70 \\
TinyLlama & 33.79 & 68.01 & 46.52 & 43.90 & 74.16 & 93.60 & 22.28 & 46.99 & 53.66 \\
\midrule
\multicolumn{10}{l}{\textit{Multilingual LLMs}} \\
mGPT & 21.93 & 48.99 & 30.65 & 48.70 & 64.53 & 63.10 & 23.26 & 41.81 & 42.87 \\
BLOOM & 24.91 & 54.59 & 34.70 & 49.75 & 67.85 & 89.30 & 25.58 & 47.39 & 49.26 \\
CroissantLLM & 26.37 & 62.58 & 40.88 & 50.00 & 72.69 & 92.70 & 23.62 & 42.89 & 51.47 \\
EuroLLM & 36.95 & 71.59 & 44.77 & 47.10 & 73.50 & 94.80 & 23.50 & 48.88 & 55.14 \\
\underline{Llama3.2} & 34.64 & 69.07 & 48.29 & 47.45 & 75.63 & 95.20 & 23.62 & 48.55 & 55.31 \\
\underline{Qwen2} & 40.02 & 72.85 & 49.17 & 39.35 & 75.79 & 96.00 & 28.76 & 48.39 & 56.29 \\
\underline{Gemma2} & 47.70 & 77.31 & 52.89 & 41.40 & 76.82 & 96.80 & 21.79 & 49.20 & 57.99 \\
\midrule
\themodel & 38.23 & 72.31 & 41.42 & 49.75 & 70.89 & 93.00 & 24.11 & 46.91 & 54.58 \\
\bottomrule
\end{tabular}
}
\caption{\thetest-English Performance.}
\label{tab:cuatrobench_english_results}
\end{table}

\subsection{Baseline models}
We benchmark \themodel{}, our pretrained model trained with \thedata{}, against an array of open-source multilingual and monolingual LLMs with varying parameter sizes, pretraining mulitlingual mixtures, and pretraining data sources.
The multilingual models in our comparison include \nameit{CroissantLLM} ($1.3$B) \citep{faysse2024croissantllm}, \nameit{Llama3.2} ($1$B) \citep{dubey2024llama}, \nameit{EuroLLM} ($1.7$B) \citep{martins2024eurollm}, \nameit{Qwen2} ($1.5$B) \citep{yang2024qwen2}, \nameit{BLOOM} ($1.1$B) \citep{le2023bloom}, \nameit{mGPT} ($1.3$B) \citep{shliazhko2022mgpt}, and \nameit{Gemma2} ($2$B) \citep{team2024gemma}. For monolingual LLMs, we include a French-centric model, \nameit{GPT-fr} ($1$B) \citep{simoulin2021modele}, and two English-centric models, \nameit{TinyLlaMA} ($1.1$B) \citep{zhang2024tinyllama} and \nameit{Pythia} ($1.4$B) \citep{biderman2023pythia}. An overview of baseline models and our \themodel{} has been shown in Table~\ref{tab:model_comparison}.

\subsection{Main results on \thetest{}}
Tables~\ref{tab:cuatrobench_french_results},~\ref{tab:cuatrobench_german_results} and~\ref{tab:cuatrobench_spanish_results} present the performance of \themodel{} compared to baseline models in French, German and Spanish. Generally, \themodel{} consistently ranks among the top three systems for French, German, and Spanish, as evidenced by the average accuracy across \thetest{} benchmarks in these languages. This performance indicates that \themodel{} significantly outperforms multilingual LLMs such as \nameit{CroissantLLM}, \nameit{BLOOM}, and \nameit{mGPT}, while achieving comparable results to strong baseline multilingual LLMs including \nameit{Gemma2}, \nameit{EuroLLM}, \nameit{Llama3.2}, and \nameit{Qwen2}, despite these latter models having comparable or slightly larger model sizes. More remarkably, \themodel{} achieves this performance while being trained on a substantially smaller number of tokens. For example, the number of tokens used to train \themodel{} is approximately $25\%$ of that used for \nameit{Gemma2} and $6\%$ of that used for \nameit{Llama3.2}, as shown in Table \ref{tab:model_comparison}.

For English, as shown in Table \ref{tab:cuatrobench_english_results}, \themodel{} demonstrates superior performance on English benchmarks compared to \nameit{TinyLlaMA}, \nameit{Pythia}, and \nameit{CroissantLLM}. This result is particularly significant given that both \nameit{TinyLlaMA} and \nameit{CroissantLLM} report training on 3 trillion tokens, a substantially larger dataset than that used for \themodel{}.


\section{Pretraining Data Analysis}
To explore the impact of various data sources, including general web data, our \thedata{}, and special data typically used during the cooldown period, we conducted a series of ablation experiments.

\subsection{Analysis setups}
\subsubsection{General web data}\label{sec:rpv2}


Given that our \thedata{} dataset is largely based on educational content which is a highly specialised domain, we investigate whether multilingual reasoning capabilities can be further enhanced by incorporating general web data.
For this purpose, we use the English, French, German, and Spanish subsets of the \nameit{RedPajama-v2}~(RPv2)~\cite{together2023redpajama} as web data. Given that web data is inherently noisy, we make further use of the quality signals provided for RPv2 and filter each subset down to a smaller, high-quality subset. Specifically, we use the six most recent dumps from 2022 and 2023 and apply quality filtering using the Gopher rules~\cite{rae2021scaling}. Additionally, web data often contains near duplicates, stemming from boilerplate text, ads, and other computer-generated text that only differs by a few words, and removing these has been shown to positively affect training efficiency and reduce the amount of memorization~\cite{lee2021deduplicating}. We therefore adopt the MinHash algorithm with locality-sensitive hashing~\cite{broder1997resemblance} to perform near-deduplication. We identify documents as near duplicates if their Jaccard similarity is greater than $0.8$ and use $128$ hash functions. 

To create our final general web dataset, we randomly sample equal numbers of tokens for English, French, German, and Spanish from the quality-filtered RPv2, ensuring that the token counts match those of each language component in \thedata{}, as detailed in Table~\ref{tab:data_statistics}. We then combine the general web data with \thedata{} with an approximately $1:1$ ratio for continued pretraining. Extending \themodel{}, we additionally pretrain for $44,000$ steps, resulting in approximately $90$ billion tokens being processed during this extended training phase. We call this continued pretrained model as \textbf{\themodelweb{}}.




\begin{table}[!t]
\centering
\resizebox{\columnwidth}{!}{%
\begin{tabular}{lccc}
\toprule
\textbf{Model} & \textbf{\# tokens} & \textbf{Method} & \textbf{Data} \\
\midrule
\themodelrealonly{} & 515B  & Train from scratch & {\makecell{\thedata{}-English \\+ RPv2}}\\
\midrule
\themodel & 515B   &  Train from scratch & \thedata\\
\midrule
\themodelweb{} & +90B &  {\makecell{Continue train \\on \themodel}} & \thedata{} + RPV2 \\
\midrule
\themodelcool{} & +20B & {\makecell{Continue train \\ on \themodelweb{}}} & {\makecell{\thedata{} \\+ RPV2 + Cooldown data}}\\
\bottomrule
\end{tabular}
}
\caption{Models used for the Pretraining Data Analysis.}
\label{tab:data-impact-model-details}
\end{table}

\subsubsection{\thedata{}}
To further investigate the impact of \thedata{} compared to general web data for pretraining, we introduce a baseline model designated as \textbf{\themodelrealonly{}}. This model is pretrained from scratch using identical hyperparameter settings, but solely with web data. The training corpus for \themodelrealonly{} comprises the English portion of \thedata{} and the filtered multilingual \nameit{RedPajama-v2} dataset used by \themodelweb{} as detailed in Section~\ref{sec:rpv2}. 

\subsubsection{Cooldown data}

As outlined in previous work~\cite{faysse2024croissantllm,zhang2024tinyllama,hu2024minicpm,martins2024eurollm}, a cooldown phase is essential for enhancing model capabilities.
Our \thedata{} dataset is comprised of high-quality educational data but lacks code and special data types such as question-answering~(QA) compared to other models. To address this, we incorporate these datasets during our cooldown stage. Specifically, we use:

\textbf{Python-Edu}~\cite{benallal2024smollmcorpus}, a dataset containing educational Python samples from The Stack, comprising 4B tokens.

\textbf{WebInstruct}~\citep{yue2024mammoth2}, a high-quality QA dataset collected from the Internet and refined through paraphrasing, consisting of $0.8$B tokens.

\textbf{WebInstruct-French}, a translated version of \nameit{WebInstruct}, consisting of $0.5$B tokens. Since our method of curating multilingual pretraining sets can scale to arbitrary high-quality datasets, we applied our translation pipeline to convert \nameit{WebInstruct} into a French version using the same Mistral model used to create \thedata{}.

For the cooldown phase, we mix this cooldown data with data from the previous stage and then train on this new data mix~(detailed in Table~\ref{tab:data-impact-model-details}) for an additional $20$B tokens with smaller learning rate.\footnote{We use constant learning rate schedule. The learning rate is $6\times 10^{-4}$ for pretraining and $6\times 10^{-5}$ for the cooldown stage.}
It is worth noting that the cooldown data represents only a small proportion of the whole corpus. The trained tokens originating from the cooldown data amount to less than $1$B, accounting for approximately $0.2\%$ of the total training tokens. We call this cooldown model as \textbf{\themodelcool}.

\subsection{Pretraining data analysis findings}
\label{sec:data_impact_finding}
We compare models with different data setups, as illustrated in Table~\ref{tab:data-impact-model-details}.


\paragraph{General web data is helpful for QA and summarization tasks.} We use FrenchBench~\citep{faysse2024croissantllm}, a benchmark that consists of QA and summarization tasks, to evaluate model performance. When comparing \themodel{} and \themodelweb{}, as shown in Table~\ref{tab:frenchbench_results}, including general web data for continued pretraining significantly improves performance with an average improvement of $2$ points ($34.37-32.41$). For the summarization task, the OrangeSum ROUGE-1 score improves from $15.35$ to $22.28$ with additional general web data, which further surpasses CroissantLLM's performance by $10+$ ROUGE-1 score. For reasoning and language understanding tasks~(Table~\ref{tab:data_impact_agg_mc}), the improvement is not as significant as for QA benchmarks. This suggests that \themodel{} is sufficiently capable of complex reasoning. The web data improves language formatting and styling, which is important for ROUGE-based metrics.

\begin{table}
\small
\centering
\resizebox{\columnwidth}{!}{%
\begin{tabular}{lccccc|c}
\toprule
\multirow{2}{*}{\textbf{Model}} & \textbf{FQuAD} & \textbf{FQuAD} & \textbf{Multi} & \textbf{Orange} & \multirow{2}{*}{\textbf{FTrivia}} & \multirow{2}{*}{\textbf{Mean}} \\
  & \textbf{GenQ} & \textbf{GenAns} & \textbf{FQuAD} & \textbf{Sum} & & \\
\midrule

\nameit{GPT-fr}  & 3.78 & 2.30 & 5.40 & 10.93 & 0.12 & 4.51 \\
\nameit{CroissantLLM}  & 18.92 & 40.50 & 32.92 & 10.36 & 51.74 & 30.89 \\
\themodelrealonly & 19.74 & 34.50 & 26.75 & 17.65 & 39.14 & 27.56 \\
\themodel  & 19.36 & 46.86 & 37.36 & 15.35 & 43.11 & 32.41 \\
\themodelweb & 20.55 & 45.32 & 36.97 & 22.28 & 46.74 & 34.37 \\
\themodelcool & \textbf{21.59} & \textbf{51.17} & \textbf{44.26} & \textbf{23.07} & \textbf{51.33} & \textbf{38.28} \\
\bottomrule
\end{tabular}
}
\caption{FrenchBench QA and Summarization Performance.}\label{tab:frenchbench_results}
\end{table}

\begin{table}
\small
\centering
\resizebox{\columnwidth}{!}{%
\begin{tabular}{lcccc}
\toprule
 & \multicolumn{4}{c}{\textbf{\thetest{}}} \\
\cmidrule{2-5}
 \textbf{Model} & \textbf{English} & \textbf{French} & \textbf{German} & \textbf{Spanish} \\
\midrule
\nameit{Gemma2}  & \textbf{57.99}  & 39.89  & 36.86 & 38.84 \\
\nameit{EuroLLM}  & 55.14  & 39.42  & 37.54 & 38.98 \\
\nameit{Llama3.2}  &  55.31 &  37.62 & 36.12 & 37.58 \\
\nameit{Qwen2}  & 56.29  & 38.05  & 35.24 & 37.62 \\
\midrule
\themodelrealonly & 50.47  & 36.81  & 36.11 & 36.61 \\
\themodel &  54.58 & 38.53  & 37.58 & 38.50 \\
\themodelweb & 53.65  & 39.88  & 36.89 & 38.74 \\
\themodelcool & 55.07 & \textbf{40.52}  & \textbf{38.52} & \textbf{39.59} \\
\bottomrule
\end{tabular}
}
\caption{Aggregated scores on \thetest, with per-dataset details provided in Table~\ref{tab:data_impact_mc_result} of Appendix~\ref{sec:data_impact_cuatroben_result}.}
\label{tab:data_impact_agg_mc}
\end{table}


\begin{table}[!t]
\small
\centering
\resizebox{1.0\columnwidth}{!}{%
\begin{tabular}{lcccc|c}
\toprule
\multirow{2}{*}{\textbf{Model}} & \multicolumn{2}{c}{\textbf{WMT14}} & \multicolumn{2}{c|}{\textbf{WMT16}} & \multirow{2}{*}{\textbf{Mean}} \\
\cmidrule(lr){2-3} \cmidrule(lr){4-5}
& \textbf{EN-FR} & \textbf{FR-EN} & \textbf{DE-EN} & \textbf{EN-DE} & \\
\midrule
\multicolumn{6}{l}{\textit{Unsupervised MT Models}}  \\
\citet{lample2017unsupervised} & 15.05 & 14.31 & 13.33 & 9.64 & 13.08 \\
\citet{artetxe2017unsupervised} & 15.13 & 15.56 & - & - & - \\
\midrule
\themodelrealonly{} & 21.69 & 23.33 & 21.67 & 17.39 & 21.02\\
\themodel{} & 24.73 & 28.76 & 28.06 & 21.14 & 25.67 \\
\themodelweb{} & 28.17 & 29.12 & 29.01 & 24.06 & 27.59 \\
\themodelcool{} & \textbf{30.88} & \textbf{32.10} & \textbf{32.81} & \textbf{26.45} & \textbf{30.56} \\

\bottomrule
\end{tabular}
}
\caption{Translation performance, reported as BLEU~\citep{papineni2002bleu}.}
\label{tab:wmt_results}
\end{table}
\paragraph{The machine-translated corpus is important for LLM performance.} To further understand the importance of pretraining on a translated corpus, we fully removed our \thedata{} dataset from the pretraining setup, keeping only general web data, resulting in \themodelrealonly{}~(detailed in Table~\ref{tab:data-impact-model-details}). As shown in Table~\ref{tab:data_impact_agg_mc}, \themodelrealonly{} consistently performs worse than \themodel{} for all languages. This performance gap is even larger for French QA benchmarks~(Table~\ref{tab:frenchbench_results}), with an average absolute difference of $4.8$ points~($32.41-27.56$). All these findings indicate that our proposed \thedata{} dataset plays a crucial role in boosting multilingual performance.

\paragraph{\themodel{} shows significant improvement using only a small amount of cooldown data.}
To our surprise, with less than $0.2\%$ of tokens pretrained from the cooldown dataset, the \themodelcool{} performance improves significantly, setting new state-of-the-art results for \thetest{}~(Table~\ref{tab:data_impact_agg_mc}). This suggests that the model trained on our \thedata{} dataset serves as a strong base model that can be adapted to downstream tasks with minimal amounts of additional data.






\section{Ablations and Discussions}
In this section, we present additional ablation experiments and provide discussions of our findings.

\subsection{Translation evaluation}
Translation tasks offer a means to investigate whether cross-lingual alignment emerges during pretraining. Unlike approaches such as \nameit{EuroLLM}~\citep{martins2024eurollm} and \nameit{CroissantLLM}~\citep{faysse2024croissantllm}, our pretraining data does not include task-specific parallel corpora~(e.g., concatenations of parallel sentence or document pairs). Thus, it is essential to assess if cross-lingual alignment occurs in our unsupervised training setup for translation. Hence, we evaluate \themodel{} on the following test sets: WMT 2014 English-French~(EN-FR and FR-EN)~\citep{bojar2014findings}, WMT 2016 English-German~(EN-DE and DE-EN)~\citep{bojar2016findings}. We use BLEU~\citep{papineni2002bleu} for evaluation. For a comparison, we benchmark \themodel{} against LLMs that, like ours, do not specifically include parallel corpora during their pretraining. Thus, we compare \themodel{} with \themodelrealonly{}, \themodelweb{}, and \themodelcool{}. Furthermore, we include conventional unsupervised machine translation~(MT) approaches~\citep{artetxe2017unsupervised,lample2017unsupervised} in our comparative analysis. 

Table~\ref{tab:wmt_results} presents the results for translation. We observe that all LLMs significantly outperform classic unsupervised MT models. Our balanced \thedata{} enables \themodel{} to achieve better cross-lingual alignment than \themodelrealonly{}. This outcome suggests that our unsupervised training approach for translation, based on a balanced multilingual dataset, demonstrates potential in addressing the challenges of unsupervised machine translation. Moreover, continuing pretraining on \themodel{} with general web data~(\themodelweb{}) significantly enhances translation performance. This improvement is further amplified by incorporating additional cooldown data. These additions substantially improve our model's in-context learning capabilities for translation tasks.

\subsection{Probing language model prior}



\begin{table}[t]
\label{tab:english-dominance}
\centering
\resizebox{1.0\columnwidth}{!}{%
\begin{tabular}{l|c|c|c|c|c}
\toprule
\textbf{Model} & \textbf{EN} (\%) & \textbf{FR} (\%) & \textbf{DE} (\%) & \textbf{ES} (\%) & \textbf{Others} (\%) \\
\midrule
\nameit{Gemma2} & 80.5 & 0.7 & 0.7 & 0.1 & 18.0 \\
\nameit{Llama3.2} & 98.8 & 0.1 & 0.0 & 0.0 & 1.1 \\
\nameit{CroissantLLM} & 37.9 & 60.4 & 0.0 & 0.0 & 1.7 \\
\nameit{Qwen2} & 47.3 & 0.2 & 0.6 & 0.4 & 51.5 \\
\nameit{EuroLLM} & 37.7 & 3.9 & 5.3 & 6.3 & 46.8 \\
\midrule
\themodel & 24.6 & 23.4 & 14.3 & 36.9 & 0.8 \\
\bottomrule
\end{tabular}
}
\caption{Language ratio on sampled texts.} 
\label{tab:english-dominance}
\end{table}


\begin{table}
\small
\centering
\resizebox{1.0\columnwidth}{!}{%
\begin{tabular}{l|ccc}
\toprule
 & \multicolumn{3}{c}{\textbf{\thetest{}}} \\
\cmidrule{2-4}
 \textbf{Model} & \textbf{English} & \textbf{French} & \textbf{German} \\
\midrule
\nameit{English Model}  & 54.56  & -  & -  \\
\nameit{English and French Model}  & 54.86  & 39.80  & - \\
\nameit{English, French, and German Model}  & 54.39  & 38.59  & 37.09  \\
\themodel{} (with all four languages)  &  54.58 & 38.53  & 37.58  \\
\bottomrule
\end{tabular}
}
\caption{``Curse of Multilinguality'' assessment.}
\label{tab:curse}
\end{table}

In this section, we analyze the prior of the pretrained language models by charting their direct outputs---the ones given by the prompt begin of sequence token---across a spectrum of languages. Specifically, we sampled $512$ generations, each with a maximum length of $300$ tokens and using a temperature of $1.0$; we then determined  the language of each generation using FastText.\footnote{\href{https://hf.co/facebook/fasttext-language-identification}{\texttt{hf.co/facebook/fasttext-language-identification}}}
We measured the prior mass of each language by calculating the percentages of generations with their corresponding labels. 
Table~\ref{tab:english-dominance} presents the results for \nameit{EuroLLM}, \nameit{Gemma2}, \nameit{Llama3.2}, and our \themodel{}.
These models are trained with different data volumes and mixtures across languages/sources.  



\paragraph{Prior mass across languages} 

Comparing the ratios in Table \ref{tab:english-dominance} gives us signals on the differences between prior masses occupied via different languages. We can see that both \textit{Gemma 2} and \nameit{Llama3.2} have extremely heavy English prior Masses, reflecting their strong biases towards English, likely due to their English-dominated pretraining corpus. In contrast, \nameit{EuroLLM}, \nameit{Qwen2} and \nameit{CroissantLLM} presents a more balanced distribution across languages, with an English ratio of $37.7$\%, $47.3$\% and $37.9$\%, respectively. This lower English dominance is indicative of a training process that may have integrated a more diverse set of languages. Our model (\themodel) demonstrates a further reduction of English prior mass with ratios of only $24.6$\%. 
Notably, \themodel{} shows significant prior mass for French ($23.4$\%), German ($14.3$\%), and Spanish ($36.9$\%), highlighting a capacity to handle these languages alongside English.
This is more likely due to our more evenly distributed pretraining data compare to the dataset used by \nameit{EuroLLM}. Such balance is crucial for LLM applications where equitable knowledge across multiple languages is required.

\paragraph{Preference of generation formats} 
While \nameit{CroissantLLM} and \nameit{EuroLLM} show a fairly balanced multilingual prior, our analysis reveals a significant issue: their generations often mix multiple languages. For instance, over \textbf{$80$\%} of \nameit{EuroLLM} generations consist of translation pairs involving two different languages (see examples in Appendix~\ref{sec:appendix-case-study}), which is high compared to the relative low proportion of translation pairs in \nameit{EuroLLM}'s pretraining data.
We hypothesize that pretraining with translation pairs led to this unexpected preference towards translation-formatted generations. 
This may be problematic as the LLM might degenerate into a translation pair generation model, thus lacking the expected diversity in natural language generation. 
On the contrary, \themodel{} excels in monolingual generation while likely maintaining a balanced multilingual prior. This makes it better suited for diverse tasks, with potentially lower language confusion~\citep{marchisio2024understanding}.
\subsection{Assessing the ``curse of multilinguality''}
\citet{conneau2019unsupervised} has demonstrated that pre-training a model with fixed capacity on an increasing number of languages improves its cross-lingual performance only up to a certain point. Beyond this threshold, performance degradation can be observed---a phenomenon known as the ``curse of multilinguality''.  In our context, it is interesting to evaluate whether this phenomenon occurs as we incrementally incorporate French, German, and Spanish components into the English component of \thedata{}. To investigate this, we compare four models pretrained from scratch: an English-only model, a bilingual model~(En and Fr), a trilingual model~(En, Fr, and De), and \themodel{} itself~(four languages). All language components for these models are sourced from \thedata{}, ensuring a consistent data distribution for our analysis. All models are trained on $515$B tokens, ensuring consistent training scale across experiments.

We evaluate the four models on the English, French, and German benchmarks from \thetest{}. Table~\ref{tab:curse} presents the average model performance across different languages. Our analysis shows no significant signs of the ``curse of multilinguality'' among these languages. Model performance on English, French, and German remain stable, with no significant performance drop when adding new languages. 

\section{Conclusion}
We introduce \thedata{}, a multilingual dataset created by machine-translating from English source texts. 
Our \themodel{} model, trained on this data, performs competitively across languages with balanced distribution.
This approach offers a scalable method for high-quality multilingual pretraining data creation, advancing multilingual research.

\section*{Limitations}
Our study yields promising results while also identifying areas for future exploration. We focused on four major European languages: English, French, German, and Spanish, which provided a solid foundation for our research. However, expanding this language set in future work could offer valuable insights into how our approach performs across a broader spectrum of linguistic diversity. This expansion could include languages with different structural properties or varying levels of resource availability, potentially encompassing low-resource languages and those from non-Indo-European families. Additionally, our experiments centered on \themodel{}, a model with $1.3$B parameters, which has demonstrated encouraging results at its current scale. However, it remains uncertain whether the benefits observed from our translated pretraining data would persist or potentially amplify when scaled to substantially larger models, such as those with $70$B+ parameters. Those larger models could potentially reveal new insights into multilingual learning dynamics and data utilization. By addressing these aspects in future research, we aim to further validate and potentially enhance the applicability and scalability of our multilingual pretraining method using machine-translated corpora. This ongoing work will contribute to the broader goal of developing more inclusive and capable multilingual language models.

\section*{Acknowledgments}
The authors acknowledge the use of computing resources provided by the Isambard-AI National AI Research Resource (AIRR). Isambard-AI is operated by the University of Bristol and is funded by the UK Government’s Department for Science, Innovation and Technology (DSIT) via UK Research and Innovation; and the Science and Technology Facilities Council [ST/AIRR/I-A-I/1023].

\bibliography{custom}

\appendix
\section{Translation Pipeline}
\begin{figure*}[!t]
\includegraphics[width=1.0\textwidth]{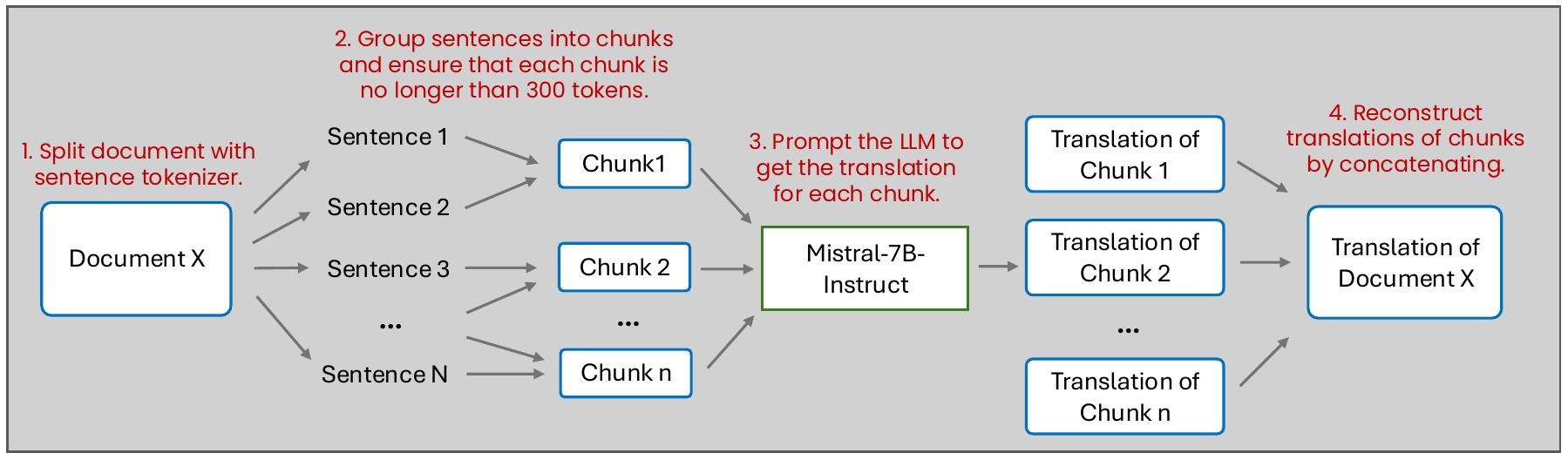}
\caption{Step-by-step illustration of the translation pipeline.}
\label{fig:chunking}
\end{figure*}

The detailed translation pipeline to produce \thedata{} is shown in Figure~\ref{fig:chunking}.
\label{sec:chunking}

\section{Hyperparameters Settings of Model Pretrainning}
\label{sec:appendix-model-arch-params}
\begin{table}[!t]
\small 
\centering
\setlength{\tabcolsep}{8pt}
\begin{tabular}{l|c}
\toprule
 Hyperparameter &  Value\\
\midrule
Sequence Length &   2048 \\
    Number of Layers &    24 \\
   Embedding Size &    2048 \\
   FFN Hidden Size &   5504 \\
   Number of Heads &   16 \\
   Position Encodings &   RoPE \\
   Activation Function &   SwiGLU\\
   Layer Norm &   RMSNorm \\
   Learning Rate &   6E-4 \\
   Batch Size &   1024 \\
   Vocabulary Size &    32000 \\
 \midrule
   Embedding Parameters &   0.13B \\
   Non-Embedding Parameters &   1.21B\\
   Total Parameters &   1.34B\\
\bottomrule
\end{tabular}
\caption{Model and pretraining hyperparameters.} 
\label{tab:model-arch}
\end{table}

Pretraining hyperparameter settings are shown in Table~\ref{tab:model-arch}.

\section{Detailed Results of Pretraining Data Analysis on \thetest{}}
\label{sec:data_impact_cuatroben_result}
Detailed \thetest{} results of models analyzed in the Pretraining Data Analysis (Section~\ref{sec:data_impact_finding}) are shown in Table~\ref{tab:data_impact_mc_result}.

\section{Generating Documents from Language Models}\label{sec:appendix-case-study}

We provide examples generated by \nameit{EuroLLM} and \nameit{CroissantLLM} in Table~\ref{tab:case-study}.
\begin{table*}[!t]
\small
\centering
\resizebox{2.0\columnwidth}{!}{%

\begin{tabular}{p{0.3\columnwidth}|p{1.4\columnwidth}|p{0.3\columnwidth}}
\toprule
Model &\textbf{Generated Example} & \textbf{Comment} \\
\midrule
EuroLLM &English: The hotel features free parking, a baggage storage and a vending machine.

Czech: Hotel nabízí parkování zdarma a prodejní stroj a obchodní centrum i tiskárny a faxovací linku. &  machine-translation pair\\ 
\midrule
EuroLLM & English: Setting: Ideal for relaxing walks and for sunbathing

German: Umgebung: Ideal zum Erfrischen oder Sonntagsspazierengehen &   machine-translation pair\\
\midrule
CroissantLLM & Un site Web sur l'électromécanique pour les services<tab>A Web site on electromechanics for services &   machine-translation pair\\
\midrule
CroissantLLM & The Moto Mods system requires Android 7.0 or higher device.<tab>* Système de Moto Mods nécessite un appareil Android 7.0 ou supérieur. &   machine-translation pair\\
\bottomrule
\end{tabular}
}
\caption{Examples of generated documents with dominant formats}
\label{tab:case-study}
\end{table*}

\begin{table*}[!t]
\begin{center}
\centering
\setlength\tabcolsep{2.5pt}
\scalebox{0.8}{
\begin{tabular}{lcccccccc}
\toprule
\textbf{Task} & Gemma2 & EuroLLM & Llama3.2 & Qwen2 & \themodelrealonly{} & \themodel{} & \themodelweb{} & \themodelcool{} \\
\midrule
ARC-C (fr)	&	35.76	&	32.34	&	27.46	&	29.60	&	25.49	&	33.53	&	33.53	&	35.93	\\
Hellaswag (fr)	&	39.70	&	40.27	&	36.02	&	38.78	&	38.32	&	38.00	&	40.15	&	41.67	\\
PAWS (fr)	&	48.55	&	51.90	&	52.20	&	48.15	&	50.20	&	52.10	&	52.00	&	51.35	\\
TruthfulQA (fr)	&	28.08	&	27.45	&	28.46	&	29.10	&	25.79	&	26.43	&	27.32	&	27.45	\\
XNLI (fr)	&	47.35	&	45.14	&	43.94	&	44.62	&	44.26	&	42.61	&	46.39	&	46.18	\\
\midrule
Average (fr)	&	39.89	&	39.42	&	37.62	&	38.05	&	36.81	&	38.53	&	39.88	&	40.52	\\
\midrule
\midrule
ARC-C (de)	&	31.14	&	29.43	&	26.78	&	26.69	&	24.89	&	31.39	&	30.97	&	33.79	\\
Hellaswag (de)	&	37.31	&	37.47	&	34.11	&	34.77	&	35.67	&	35.70	&	37.43	&	39.39	\\
PAWS (de)	&	42.55	&	46.00	&	47.95	&	43.40	&	50.45	&	51.35	&	45.35	&	48.65	\\
TruthfulQA (de)	&	26.65	&	28.68	&	27.54	&	28.55	&	26.27	&	26.90	&	25.25	&	25.13	\\
XNLI (de)	&	46.63	&	46.10	&	44.22	&	42.77	&	43.25	&	42.57	&	45.46	&	45.66	\\
\midrule
Average (de)	&	36.86	&	37.54	&	36.12	&	35.24	&	36.11	&	37.58	&	36.89	&	38.52	\\
\midrule
\midrule
ARC-C (es)	&	36.15	&	32.91	&	28.72	&	30.51	&	27.01	&	32.99	&	32.31	&	35.73	\\
Hellaswag (es)	&	41.69	&	41.00	&	37.13	&	39.03	&	37.98	&	38.66	&	40.14	&	42.13	\\
PAWS (es)	&	44.00	&	49.75	&	51.55	&	44.05	&	50.25	&	50.00	&	48.80	&	47.90	\\
TruthfulQA (es)	&	28.01	&	27.25	&	27.50	&	31.43	&	24.59	&	27.88	&	27.63	&	27.76	\\
XNLI (es)	&	44.34	&	43.98	&	43.01	&	43.09	&	43.21	&	42.97	&	44.82	&	44.42	\\
\midrule
Average (es)	&	38.84	&	38.98	&	37.58	&	37.62	&	36.61	&	38.50	&	38.74	&	39.59	\\
\midrule
\midrule
ARC-C (en)	&	47.70	&	36.95	&	34.64	&	40.02	&	27.65	&	38.23	&	36.18	&	39.51	\\
ARC-E (en)	&	77.31	&	71.59	&	69.07	&	72.85	&	61.49	&	72.31	&	70.08	&	72.85	\\
Hellaswag (en)	&	52.89	&	44.77	&	48.29	&	49.17	&	38.53	&	41.42	&	41.23	&	43.93	\\
PAWS (en)	&	41.40	&	47.10	&	47.45	&	39.35	&	47.55	&	49.75	&	47.70	&	47.30	\\
PIQA (en)	&	76.82	&	73.50	&	75.63	&	75.79	&	68.28	&	70.89	&	71.22	&	72.47	\\
SciQ (en)	&	96.80	&	94.80	&	95.20	&	96.00	&	91.00	&	93.00	&	92.90	&	94.20	\\
TruthfulQA (en)	&	21.79	&	23.50	&	23.62	&	28.76	&	22.28	&	24.11	&	23.38	&	22.40	\\
XNLI (en)	&	49.20	&	48.88	&	48.55	&	48.39	&	46.99	&	46.91	&	46.47	&	47.91 \\
\midrule
Average (en)	&	57.99	&	55.14	&	55.31	&	56.29	&	50.47	&	54.58	&	53.65	&	55.07	\\
\bottomrule
\end{tabular}
}
\end{center}
\caption{Results of Data Impact Analysis on \thetest{}, reported with the Accuracy metric.}
\label{tab:data_impact_mc_result}
\end{table*}

\section{Language Variance}
We observe that \themodel{} demonstrates consistent performance for different types of benchmarks across all four languages when we analyze the average model performance for each specific benchmark type across the four languages. As shown in Table~\ref{tab:task_std}, compared to multilingual LLM baselines such as \nameit{Gemma2}, \nameit{Llama3.2}, \nameit{EuroLLM}, and \nameit{Qwen2}, \themodel{} exhibits the lowest average standard deviations in model performance across different benchmarks. This suggests that the balanced language distribution of \thedata{} that \themodel{} uses potentially contributes to bridging languages performance gaps for multilingual pretraining.

\begin{table}[!t]
\centering
\resizebox{\columnwidth}{!}{%
\begin{tabular}{l|c|cccc}
\toprule
& \multicolumn{1}{c|}{\multirow{2}{*}{\textbf{\themodel{}}}} & \multicolumn{4}{c}{\textbf{Baselines}} \\
\cmidrule{3-6}
\textbf{Task} &  & Gemma2 & EuroLLM & Llama3.2 & Qwen2 \\
\midrule
ARC-C	&	\textbf{2.94}	&	7.05	&	3.10	&	3.58	&	5.78 \\
Hellaswag	&	\textbf{2.35}	&	6.90	&	3.01	&	6.39	&	6.14 \\
PAWS	&	\textbf{1.12}	&	3.14	&	2.66	&	2.43	&	3.60 \\
TruthfulQA	&	1.60	&	2.97	&	2.24	&	2.15	&	\textbf{1.33} \\
XNLI &	2.10	&	\textbf{2.01}	&	2.09	&	2.47	&	2.58 \\
\midrule
Average &	\textbf{2.02}	&	4.41	&	2.62	&	3.41	&	3.89 \\
\bottomrule
\end{tabular}
}
\caption{Standard deviation of model performance across languages.}
\label{tab:task_std}
\end{table}

\end{document}